\documentclass[runningheads]{llncs}
\usepackage[utf8]{inputenc} 
\usepackage{graphicx}
\usepackage{hyperref}

\usepackage[dvipsnames]{xcolor}
\usepackage{amsmath,amsfonts,amssymb}
\usepackage{subcaption}
\usepackage{multirow}
\usepackage{booktabs}
\usepackage[misc]{ifsym}

\makeatletter
\newcommand{\printfnsymbol}[1]{%
  \textsuperscript{\@fnsymbol{#1}}%
}

\begin{document}
\title{ISINet: An Instance-Based Approach for Surgical Instrument Segmentation}
%
%
\author{Cristina González \inst{1}\thanks{Both authors contributed equally to this work.}\orcidID{0000-0001-9445-9952}(\Letter) \and
Laura Bravo-Sánchez\inst{1}\printfnsymbol{1}\orcidID{0000-0003-3556-3391} \and
Pablo Arbelaez\inst{1}\orcidID{0000-0001-5244-2407}}
%
\authorrunning{C. González et al.}
%
\institute{Center for Research and Formation in Artificial Intelligence, \\
Universidad de los Andes, Bogotá, Colombia \\
\email{\{ci.gonzalez10, lm.bravo10, pa.arbelaez\}@uniandes.edu.co}}
\maketitle              
%
\begin{abstract}
We study the task of semantic segmentation of surgical instruments in robotic-assisted surgery scenes. We propose the Instance-based Surgical Instrument Segmentation Network (ISINet), a method that addresses this task from an instance-based segmentation perspective. Our method includes a temporal consistency module that takes into account the previously overlooked and inherent temporal information of the problem. We validate our approach on the existing benchmark for the task, the Endoscopic Vision 2017 Robotic Instrument Segmentation Dataset \cite{endovis2017}, and on the 2018 version of the dataset \cite{EndoVis2018}, whose annotations we extended for the fine-grained version of instrument segmentation. Our results show that ISINet significantly outperforms state-of-the-art methods, with our baseline version \textbf{duplicating} the Intersection over Union (IoU) of previous methods and our complete model \textbf{triplicating} the IoU.

\end{abstract}

\section{Introduction}
In this paper, we focus on the task of semantic segmentation of surgical instruments in robotic-assisted surgery scenes. In other words, we aim at identifying the instruments in a surgical scene and at correctly labeling each instrument pixel with its class. The segmentation of surgical instruments or their type is frequently used as an intermediate task for the development of computer-assisted surgery systems \cite{cas-review} such as instrument tracking \cite{retinamethod1}, pose estimation \cite{pelvicdataset}, and surgical phase estimation \cite{workflow-challenge}, which in turn have applications ranging from operating room optimization to personalization of procedures, and particularly, in preoperative planning \cite{cas-review,endonet,workflow-review}. Hence, developing reliable methods for the semantic segmentation of surgical instruments can advance multiple fields of research.

The task of instrument segmentation in surgical scenes was first introduced in the Endoscopic Vision 2015 Instrument Segmentation and Tracking Dataset \cite{endovis2015}. However, the objective was not to distinguish among instrument types, but to extract the instruments from the background and label their parts. The dataset's annotations were obtained using a semi-automatic method, leading to a misalignment between the groundtruth and the images \cite{endovis2017}. Another limitation of this pioneering effort was the absence of substantial background changes, which further simplified the task. 

\begin{figure*}[ht]
\centering
\includegraphics[width=0.8\linewidth]{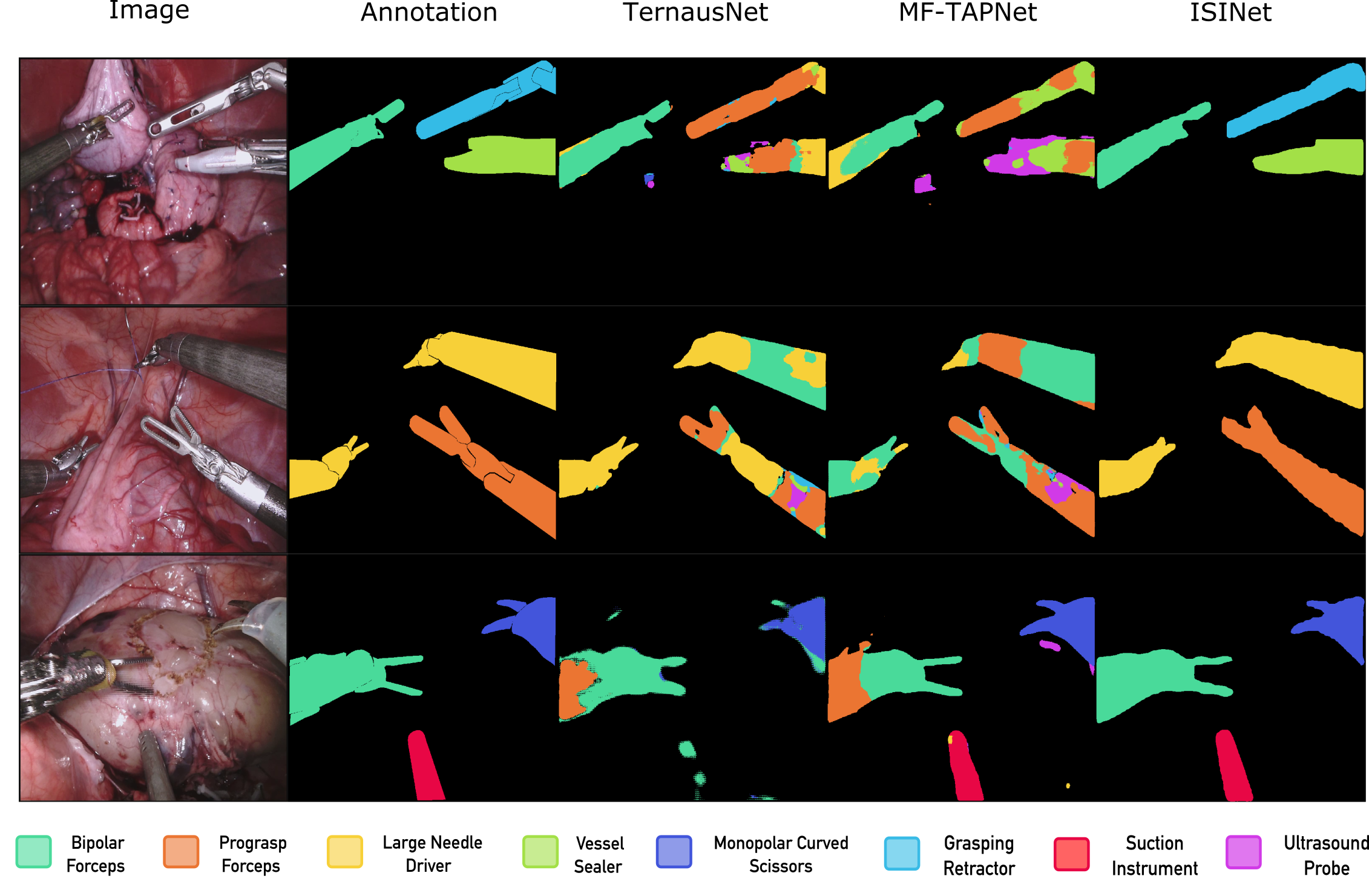}
\caption{Each row depicts an example result for the task of instrument type segmentation on the EndoVis 2017 and 2018 datasets. The columns from left to right: image, annotation, segmentation of TernausNet~\cite{ternausnet}, segmentation of MFTAPNet~\cite{mftapnet} and the segmentation of our method ISINet. The instrument colors represent instrument types. Best viewed in color.}
\label{fig:qual2017}
\end{figure*}

The Endoscopic Vision 2017 Robotic Instrument Segmentation (EndoVis 2017) Dataset~\cite{endovis2017} was developed to overcome the drawbacks of the 2015 benchmark. This dataset contains 10 robotic-assisted surgery image sequences, each composed of 225 frames. Eight sequences make up the training data and two sequences the testing data. The image sequences show up to 5 instruments per frame, pertaining to 7 instrument types. In this dataset, the task was modified to include annotations for instrument types as well as instrument parts. To date, this dataset remains the only existing experimental framework to study this fine-grained version of the instrument segmentation problem. Despite the effort put into building this dataset, it still does not reflect the general problem, mainly due to the limited amount of data, unrealistic surgery content (the videos are recorded from skills sessions), and the sparse sampling of the original videos, which limits temporal consistency.

The next installment of the problem, the Endoscopic Vision 2018 Robotic Scene Segmentation Dataset, increased the complexity of surgical image segmentation by including anatomical objects such as organs and non-robotic surgical instruments like gauze and suturing thread. In contrast to the 2017 dataset, these images were taken from surgical procedures and thus boast a large variability in backgrounds, instrument movements, angles, and scales. Despite the additional annotations, the instrument class was simplified to a general \textit{instrument} category that encompasses all instrument types. For this reason, the 2018 dataset cannot be used for the 2017 fine-grained version of the instrument segmentation task.

State-of-the-art methods for the segmentation of surgical instruments follow a pixel-wise semantic segmentation paradigm in which the class of each pixel in an image is predicted independently. Most methods~\cite{endovis2017,2015method,2018comparison,ternausnet,mftapnet} modify the neural network U-Net \cite{unet}, which in turn is based on Fully Convolutional Networks (FCN) \cite{fcn}. Some of these methods attempt to take into account details that could differentiate the instrument as a whole, by using boundaries~\cite{toolnet}, depth perception~\cite{streoscennet}, post-processing strategies~\cite{2017grabcut}, saliency maps~\cite{saliency} or pose estimation~\cite{kurmann2017}. Nevertheless, these techniques have a label consistency problem in which a single instrument can be assigned multiple instrument types, that is, there is a lack of spatial consistency in class labels within objects. \cite{kletz2019} addresses this challenge by employing an instance-based segmentation approach; however, their work on gynecological instruments was developed on a private dataset.

The second limitation of state-of-the-art models for this task is the difficulty in ensuring label consistency for an instrument through time, that is, usually the instrument classes are predicted frame by frame without considering the segmentation labels from previous frames. Recently, MF-TAPNet~\cite{mftapnet} was the first method to include a temporal prior to enhance segmentation. This prior is used as an attention mechanism and is calculated using the optical flow of previous frames. Other methods that use temporal cues have been mostly developed for surgical instrument datasets that focus on instrument tracking instead of instrument segmentation \cite{retinadataset,pelvicdataset,detectiondataset,rtmdnet}. More recently, those methods developed in \cite{robust-mis} employ temporal information for improving the segmentations or for data augmentation purposes. Instead of using temporal information to improve the segmentations, our method employs the redundancy in predictions across frames to correct mislabeled instruments, that is, to ensure temporal consistency.

In this paper, we address the label consistency problem by introducing an instance-based segmentation method for this task, the \textbf{I}nstance-based \textbf{S}urgical \textbf{I}nstrument Segmentation \textbf{Net}work (ISINet). Figure \ref{fig:qual2017} shows examples of the segmentation of robotic-assisted surgery scenes predicted by the state-of-the-art TernausNet~\cite{ternausnet} compared to the result of ISINet. In contrast to pixel-wise segmentation methods, our approach first identifies instrument candidates, and then assigns a unique category to the complete instrument. Our model builds on the influential instance segmentation system Mask R-CNN~\cite{mask}, which we adapted to the instrument classes present in the datasets, and to which we added a temporal consistency module that takes advantage of the sequential nature of the data. Our temporal consistency strategy (i) identifies instrument instances over different frames in a sequence, and (ii) takes into account the class predictions of consecutive individual frames to generate a temporally consistent class prediction for a given instance.

As mentioned above, a limiting factor in solving instrument segmentation is the relative scarcity of annotated data, particularly for the fine-grained version of the task. In order to quantitatively assess the influence of this factor in algorithmic performance, we collect additional instrument type annotations for the 2018 dataset, extending thus the 2017 training data. Following the 2017 dataset annotation protocol, we manually annotate the instruments with their types and temporally consistent instance labels, with the assistance of a specialist. Thus, with our additional annotations, we augment the data available for this task with 15 new image sequences, each composed of 149 frames, and provide annotations for new instrument types. Our annotations make the EndoVis 2018 dataset the second experimental framework available for studying the instrument segmentation task. 

We demonstrate the efficacy of our approach by evaluating ISINet's performance in both datasets, with and without using the additional data during training. In all settings, the results show that by using an instance-based approach, we \textbf{duplicate} and even \textbf{triplicate} the performance of the state-of-the-art pixel-wise segmentation methods.

Our main contributions can be summarized as follows:
\begin{enumerate}
\item We present ISINet, an instance-based method for instrument segmentation.
\item We propose a novel temporal consistency module that takes advantage of the sequential nature of data to increase the accuracy of the class predictions of instrument instances.
\item We provide new data for studying the instrument segmentation task, and empirically demonstrate the need for this additional data.
\end{enumerate}

To ensure reproducibility of our results and to promote further research on this task, we provide the pre-trained models, source code for ISINet and additional annotations created for the EndoVis 2018 dataset \footnote{\url{https://github.com/BCV-Uniandes/ISINet}}.

\section{ISINet}
Unlike pixel-wise segmentation methods, which predict a class for each pixel in the image, instance-based approaches produce a class label for entire object instances. Our method, Instance-based Surgical Instrument segmentation Network (ISINet), builds on the highly successful model for instance segmentation in natural images, Mask R-CNN. We adapt this architecture to the fine-grained instrument segmentation problem by modifying the prediction layer to the number of classes found in the EndoVis 2017 and 2018 datasets, and develop a module to promote temporal consistency in the per-instance class predictions across consecutive frames. Our temporal consistency module works in two steps: first, in the \textit{Matching Step}, for each image sequence we identify and follow the instrument instances along the sequence and then, in the \textit{Assignment Step} we consider all the predictions for each instance and assign an overall instrument type prediction for the instance. 

Initially, for the images in a sequence $I$ from frame $t=1$ to the final frame $T$, we obtain via a candidate extraction model ($M$), particularly Mask R-CNN, a set of $n$ scores ($S$), object candidates ($O$) and class predictions ($C$) for every frame $t$. Where $n$ correspond to all the predictions with a confidence score above 0.75.
\begin{equation*}
(\{\{S_{i,t}\}_{i=1}^{n}\}_{t=1}^{T},  \{\{O_{i,t}\}_{i=1}^{n}\}_{t=1}^{T},  \{\{C_{i,t}\}_{i=1}^{n}\}_{t=1}^{T}) = M(\{I_t\}_{t=1}^{T})
\end{equation*}

We calculate the backward optical flow ($OF$), that is from frame $t$ to $t-1$, for all the consecutive frames in a sequence. For this purpose we use FlowNet2~\cite{flownet2} ($F$) pre-trained on the MPI Sintel dataset \cite{sintel} and use the PyTorch implementation \cite{flownet2-implement}. 
\begin{equation*}
    OF_{{t}\rightarrow{t-1}} = F([I_t, I_{t-1}])
\end{equation*}

\textbf{Matching Step}.
For the candidate matching step given a frame $t$, we retrieve the candidates $\{\{O_{i,t}\}_{i=1}^{n}\}_{t=t-f}^{t}$, scores $\{\{S_{i,t}\}_{i=1}^{n}\}_{t=t-f}^{t}$ and class $\{\{C_{i,t}\}_{i=1}^{n}\}_{t=t-f}^{t}$ predictions from the $f$ previous frames. We use the optical flow to iteratively warp each candidate from the previous frames to the current frame $t$. For example, to warp frame $t-2$ into frame $t$, we apply the following equation from frame $t-2$ to $t-1$, and from $t-1$ to $t$.

\begin{equation*}
    \hat{O}_{i,t-1} = \underset{{t-1}\rightarrow{t}}{Warp}(O_{i,t-1}, OF_{{t}\rightarrow{t-1}})
\end{equation*}

Once we obtain the warped object candidates $\hat{O}$, we follow an instrument instance through time by matching every warped object candidate from the $f$ frames and the current frame $t$ amongst themselves by finding reciprocal pairings in terms of the Intersection over Union (IoU) metric between each possible candidate pair. Additionally, we only consider reciprocal pairings that have an IoU larger than a threshold $U$. The end result of the matching step is a set $O$ of $m$ instances along frames $t-f$ to $t$, an instance $O_k$ need not be present in all the frames.
\begin{equation*}
    \{\{O_{k,t}\}_{k=1}^{m}\}_{t=t-f}^{t} = Matching(\{\{\hat{O}_{i,t}\}_{i=1}^{n}\}_{t=t-f}^{t})
\end{equation*}

\textbf{Assignment Step}.
The objective of this step is to update the class prediction for each instance in the current frame, by considering the predictions of the previous $f$ frames. For this purpose, we use a function $A$ that considers both the classes and scores for each instance $k$. 
\begin{equation*}
    C_{k,t} = A([C_{k,t-f}, \cdots, C_{k,t}], [S_{k,t-f}, \cdots, S_{k,t}])
\end{equation*}

We repeat the Matching and Assignment steps for every frame $t$ in a sequence and for all the sequences in a set. For our final method we set $f$ to 6, $U$ to 0 and 0.5 for the 2017 and 2018 datasets respectively, and define $A$ as the mode of the input classes weighted by their scores, we validate these parameters experimentally in the supplementary material.

\textit{Implementation details}. For training Mask R-CNN we use the official implementation \cite{mask-implement}. We train until convergence with the 1x schedule of the implementation, a learning rate of 0.0025, weight decay of $1e^{-4}$, and 4 images per batch on an NVIDIA TITAN-X Pascal GPU. Additionally, for all experiments we use a ResNet-50~\cite{resnet} backbone pre-trained on the MS-COCO dataset \cite{coco}.

\section{Experiments}
\subsection{Additional Annotations for EndoVis 2018}
We provide additional instrument-type data for the task of instrument-type segmentation in surgical scenes. For this purpose, we manually extend the annotations of the EndoVis 2018 with the assistance of a specialist. Originally, this dataset's instruments are annotated as a general \textit{instrument} class and are labeled with their parts (\textit{shaft}, \textit{wrist} and \textit{jaws}). To make this dataset available for the study of fine-grained instrument segmentation, that is, to distinguish among instrument types, we further annotate each instrument in the dataset with its type. Based on the classes presented in the 2017 version of this dataset and the DaVinci systems catalog \cite{davinci}, we identify 9 instrument types: Bipolar Forceps (both Maryland and Fenestrated), Prograsp Forceps, Large Needle Driver, Monopolar Curved Scissors, Ultrasound Probe, Suction Instrument, Clip Applier, and Stapler. However, we refrain from evaluating the Stapler class due to the limited amount of examples.

For the dataset's 15 image sequences, each composed of 149 frames, we manually extract each instrument from the other objects in the scene and assign it one of the 10 aforementioned types. We label each instrument by taking into account its frame and its complete image sequence to ensure a correct label in blurry or partially occluded instances. We maintain the instrument part annotations as additional information useful for grouping-based segmentation methods. Furthermore, ensure that our instance label annotations are consistent throughout the frames of a sequence to make the dataset suitable for training instance-based segmentation methods. Our annotations are compatible with the original scene segmentation task and with the MS-COCO standard dataset format. 

\subsection{Experimental Setup}
For our experimental framework, we use the EndoVis 2018 and 2017 datasets. In both datasets the images correspond to robot-assisted surgery videos taken with the DaVinci robotic system, and the annotations are semantic and instance segmentations of robotic instrument tools.

For the experimental validation process, we divide the original training images of the 2018 dataset into two sets, the validation set with sequences 2, 5, 9, and 15, while the remaining sequences are part of the training set. As the 2017 dataset is smaller we use 4-fold cross-validation with the standard folds described in \cite{ternausnet}.
For the quantitative evaluation we use three metrics based on the Intersection over Union (IoU) metric, each more stringent than the preceding metric. For a prediction $P$ and groundtruth $G$ in a frame $i$, we compute the challenge IoU, the metric from the 2017 challenge, which only considers the classes that are present in a frame. A variation of the IoU (eq.\ref{eq:iou}) averaged over all the classes $C$ and frames $N$, and the mean class IoU (eq.\ref{eq:meanciou}) which corresponds to the IoU per class averaged across the classes.
\begin{equation} \label{eq:iou}
    IoU = \frac{1}{N} \sum_{i=1}^{N} {\left(\frac{1}{C}\sum_{c=1}^{C} \frac{P_{i,c} \cap G_{i,c}}{P_{ic} \cup G_{i,c}}\right)}
\end{equation}

\begin{equation} \label{eq:meanciou}
    mean\;cIoU = \frac{1}{C} \sum_{c=1}^{C} {\left(\frac{1}{N}\sum_{i=1}^{N} \frac{P_{i,c} \cap G_{i,c}}{P_{i,c} \cup G_{i,c}}\right)}
\end{equation}
\begin{table}[t]
\caption{Comparison of ISINet against the state-of-the-art methods for this task. The results are shown in terms of the IoU per class, mean IoU across classes, challenge IoU and IoU. D stands for use of additional data and T for temporal consistency module. Best values in bold.}
\begin{subtable}{\textwidth}
\centering
\caption{Comparison against the state-of-the-art on the EndoVis 2017 dataset.}
\label{tab:res2017}
\resizebox{\textwidth}{!}{%
\begin{tabular}{ccccccccccccc}
\toprule
\multirow{3}{*}{Method} & \multirow{3}{*}{D} & \multirow{3}{*}{T} & \multirow{3}{*}{\begin{tabular}[c]{@{}c@{}}challenge\\ IoU\end{tabular}} & \multirow{2}{*}{IoU} & \multicolumn{7}{c}{Instrument classes} & \multirow{3}{*}{\begin{tabular}[c]{@{}c@{}}mean \\ class \\ IoU\end{tabular}} \\
 \cmidrule{6-12}
 &  &  &  &  & \begin{tabular}[c]{@{}c@{}}Bipolar\\ Forceps\end{tabular} & \begin{tabular}[c]{@{}c@{}}Prograsp\\ Forceps\end{tabular} & \begin{tabular}[c]{@{}c@{}}Large\\ Needle Driver\end{tabular} & \begin{tabular}[c]{@{}c@{}}Vessel\\ Sealer\end{tabular} & \begin{tabular}[c]{@{}c@{}}Grasping\\ Retractor\end{tabular} & \begin{tabular}[c]{@{}c@{}}Monopolar \\ Curved Scissors\end{tabular} & \begin{tabular}[c]{@{}c@{}}Ultrasound \\ Probe\end{tabular} &  \\
 \midrule
\multicolumn{1}{c}{TernausNet\cite{ternausnet}} & \multicolumn{1}{c}{} & \multicolumn{1}{c}{} & 35.27 & 12.67 & 13.45 & 12.39 & 20.51 & 5.97 & 1.08 & 1.00 & \textbf{16.76} & 10.17 \\
\multicolumn{1}{c}{MF-TAPNet\cite{mftapnet}} & \multicolumn{1}{c}{} & \multicolumn{1}{c}{} & 37.35 & 13.49 & 16.39 & 14.11 & 19.01 & 8.11 & 0.31 & 4.09 & 13.40 & 10.77 \\
\multicolumn{1}{c}{\multirow{2}{*}{ISINet (Ours)}} & \multicolumn{1}{c}{} & \multicolumn{1}{c}{} & 53.55 & 49.57 & 36.93 & 37.80 & 47.06 & 24.96 & \textbf{2.01} & 19.99 & 13.90 & 26.92 \\
\multicolumn{1}{c}{} & \multicolumn{1}{c}{} & \multicolumn{1}{c}{\checkmark} & \textbf{55.62} & \textbf{52.20} & \textbf{38.70} & \textbf{38.50} & \textbf{50.09} & \textbf{27.43} & \textbf{2.01} & \textbf{28.72} & 12.56 & \textbf{28.96} \\
 \hline
\multicolumn{1}{c}{\multirow{2}{*}{ISINet (Ours)}} & \multicolumn{1}{c}{\checkmark} & \multicolumn{1}{c}{} & 66.27 & 62.70 & 59.53 & 45.73 & 58.65 & 24.38 & 2.87 & 35.85 & \textbf{28.33} & 36.48 \\
\multicolumn{1}{c}{} & \multicolumn{1}{c}{\checkmark} & \multicolumn{1}{c}{\checkmark} & \textbf{67.74} & \textbf{65.18} & \textbf{62.86} & \textbf{46.46} & \textbf{64.12} & \textbf{27.77} & \textbf{3.06} & \textbf{37.12} & 25.18 & \textbf{38.08} \\
 \bottomrule
\end{tabular}%
}
\end{subtable}

\begin{subtable}{\textwidth}
\centering
\caption{Comparison against the state-of-the-art on the EndoVis 2018 dataset.}
\label{tab:res2018}
\resizebox{\textwidth}{!}{%
\begin{tabular}{ccccccccccccc}
\toprule
\multirow{3}{*}{Method} & \multirow{3}{*}{D}  & \multirow{3}{*}{T} & \multirow{3}{*}{\begin{tabular}[c]{@{}c@{}}challenge \\ IoU\end{tabular}} & \multirow{3}{*}{IoU} & \multicolumn{7}{c}{Instrument classes}  
 & \multirow{3}{*}{\begin{tabular}[c]{@{}c@{}}mean \\ class\\  IoU \end{tabular}} 
 \\ \cmidrule{6-12}
 &  &  &  &  & \begin{tabular}[c]{@{}c@{}}Bipolar\\ Forceps\end{tabular} & \begin{tabular}[c]{@{}c@{}}Prograsp\\ Forceps\end{tabular} & \begin{tabular}[c]{@{}c@{}}Large\\ Needle Driver\end{tabular} & \begin{tabular}[c]{@{}c@{}}Monopolar\\ Curved Scissors\end{tabular} & \begin{tabular}[c]{@{}c@{}}Ultrasound\\ Probe\end{tabular} & \begin{tabular}[c]{@{}c@{}}Suction\\ Instrument\end{tabular} & \begin{tabular}[c]{@{}c@{}}Clip\\ Applier\end{tabular} &  \\
 \midrule
\multicolumn{1}{c}{TernausNet\cite{ternausnet}} & \multicolumn{1}{c}{} & \multicolumn{1}{c}{} & 46.22 & 39.87 & 44.20 & 4.67 & 0.00 & 50.44 & 0.00 & 0.00 & 0.00 & 14.19 \\
\multicolumn{1}{c}{MF-TAPNet\cite{mftapnet}} & \multicolumn{1}{c}{} & \multicolumn{1}{c}{} & 67.87 & 39.14 & 69.23 & 6.10 & 11.68 & 70.24 & 0.57 & 14.00 & \textbf{0.91} & 24.68 \\
\multicolumn{1}{c}{\multirow{2}{*}{ISINet (Ours)}} & \multicolumn{1}{c}{} & \multicolumn{1}{c}{} & 72.99 & \textbf{71.01} & 73.55 & \textbf{48.98} & 30.38 & \textbf{88.17} & \textbf{2.23} & \textbf{37.84} & 0.00 & 40.16 \\
\multicolumn{1}{c}{} & \multicolumn{1}{c}{} & \multicolumn{1}{c}{\checkmark} & \textbf{73.03} & 70.97 & \textbf{73.83} & 48.61 & \textbf{30.98} & 88.16 & 2.16 & 37.68 & 0.00 & \textbf{40.21} \\
 \cmidrule{1-13}
 \multicolumn{1}{c}{\multirow{2}{*}{ISINet (Ours)}} & \multicolumn{1}{c}{\checkmark} & \multicolumn{1}{c}{} & 77.19 & 75.25 & 76.55 & 48.79 & 50.24 & \textbf{91.50} & 0.00 & 44.95 & 0.00 & 44.58 \\
 \multicolumn{1}{c}{} & \multicolumn{1}{c}{\checkmark} & \multicolumn{1}{c}{\checkmark} & \textbf{77.47} & \textbf{75.59} & \textbf{76.60} & \textbf{51.18} & \textbf{52.31} & 91.08 & 0.00 & \textbf{45.87} & 0.00 & \textbf{45.29} \\
    \bottomrule
\end{tabular}%
}
\end{subtable}
\end{table}
\subsection{Experimental Validation}
We compare the performance of our approach ISINet with the state-of-the-art methods TernausNet \cite{ternausnet} and MF-TAPNet \cite{mftapnet}. For TernausNet we use the pre-trained models provided for the 2017 dataset, and for the 2018 dataset we retrain using the official implementation with the default 2017 parameters and a learning rate of $1e^{-4}$. For MF-TAPNet, we retrain the method for both datasets with the official implementation and the default parameter setting. We present the results of this experiment for the EndoVis 2017 dataset on Table~\ref{tab:res2017} and on Table~\ref{tab:res2018} for the EndoVis 2018 dataset.

The results show that our baseline method, that is, without the temporal consistency module nor the additional data, outperforms TernausNet and MF-TAPNet in all the overall IoU metrics in both datasets, \textbf{duplicating} or \textbf{triplicating} the performance of the state-of-the-art depending on the metric. For some instrument classes, the improvement is more than 30.0 IoU on both datasets. We observe that the improvement correlates with the number of examples of a class, as can be seen by the performance of the Grasping Retractor and Clip Applier instruments. Figure~\ref{fig:qual2017} depicts the advantages of our method over the state-of-the-art in the 2017 dataset, by segmenting previously unidentified classes and recovering complete instruments. Please refer to the supplementary material for additional qualitative results, including error modes.

We design an experiment to assess the effect of training on additional data and compare the results against only training with one dataset. For the 2018 dataset the additional data is the complete 2017 dataset. However, as the EndoVis 2017 dataset uses a 4-fold validation scheme, we train on three folds and the 2018 data, and evaluate on the remaining fold. The final result is the average of the results on all folds. Considering that not all classes are present in both datasets, we only predict segmentations for each dataset's existing classes. However, we train with all the available examples. Tables \ref{tab:res2017} and \ref{tab:res2018} demonstrate that, for both datasets, training on additional data results in better performance compared to training on a single dataset. The performance of 6 out of 7 classes on the EndoVis 2017 dataset improves, and 4 out of 7 classes of the EndoVis 2018 dataset follow this trend, with some of them increasing by up to \textbf{20.0 IoU} percentage points. These results confirm the need for additional data in order to solve this task.

In order to validate our temporal consistency module, we evaluate ISINet with and without the module (T) and with and without using additional data (D). We perform these experiments on both datasets. Tables~\ref{tab:res2017} and \ref{tab:res2018} show that our temporal consistency module improves the overall metrics with and without additional data on both datasets. Our module corrects outlier predictions in all the classes except the Ultrasound Probe, with some of them increasing nearly 8 percentage points on the 2017 dataset. Despite the overall gain on the 2018 dataset, we hypothesize that the increment is less compared to the 2017 dataset due to the higher variability of the 2018 dataset.

\section{Conclusions}
In this paper, we address the task of instrument segmentation in surgical scenes by proposing an instance-based segmentation approach. Additionally, we propose a temporal consistency module that considers an instance's predictions across the frames in a sequence. Our method ISINet outperforms the state-of-the-art semantic segmentation methods in the benchmark dataset EndoVis 2017 and the EndoVis 2018 dataset. We extend the former dataset for this task by manually annotating the instrument types. Additionally, our results indicate that using more data during the training process improved model generalization for both datasets. Finally, we observe that our temporal consistency module enhances performance by better preserving the identity of objects across time. We will provide the code and pre-trained models for both datasets, and the instrument type annotations for the EndoVis 2018 dataset.

\paragraph{Acknowledgments} The authors thank Dr. Germán Rosero for his support in the verification of the instrument type annotations.

%
%
\bibliographystyle{splncs04}
%

\title{Supplementary Material for ISINet}
%
%
\author{Cristina González \inst{1}\thanks{Both authors contributed equally to this work.}\orcidID{0000-0001-9445-9952} \and
Laura Bravo-Sánchez\inst{1}\printfnsymbol{1}\orcidID{0000-0003-3556-3391} \and
Pablo Arbelaez\inst{1}\orcidID{0000-0001-5244-2407}}
\authorrunning{C. González et al.}
%
\institute{Center for Research and Formation in Artificial Intelligence,\\
Universidad de los Andes, Bogotá, Colombia \\
\email{\{ci.gonzalez10, lm.bravo10, pa.arbelaez\}@uniandes.edu.co}}
\maketitle              

\begin{table}[h]
\caption{Ablation results of ISINet's temporal consistency module on the group 1 and 2 splits. Threshold, number of frames, and class selection method refer to the module's parameters. Most parameters improve the performance of our baseline ISINet, we select the parameters than most increase the result on both datasets.}
\begin{subtable}{\textwidth}
\centering
\caption{Results on the EndoVis 2017 dataset.}
\label{tab:res2017}
\resizebox{1.0\textwidth}{!}{%
\begin{tabular}{ccccccccccc}
\toprule
\multirow{3}{*}{Threshold} & \multirow{3}{*}{\begin{tabular}[c]{@{}c@{}} Number\\ of frames\end{tabular}} & \multirow{3}{*}{\begin{tabular}[c]{@{}c@{}} Assignment\\ Strategy\end{tabular}} & \multicolumn{7}{c}{Instrument classes} & \multirow{3}{*}{\begin{tabular}[c]{@{}c@{}}mean \\ class \\ IoU\end{tabular}} \\
\cmidrule{4-10}
&  &  & \begin{tabular}[c]{@{}c@{}}Bipolar\\ Forceps\end{tabular} & \begin{tabular}[c]{@{}c@{}}Prograsp\\ Forceps\end{tabular} & \begin{tabular}[c]{@{}c@{}}Large\\ Needle Driver\end{tabular} & \begin{tabular}[c]{@{}c@{}}Monopolar\\ Curved Scissors\end{tabular} & \begin{tabular}[c]{@{}c@{}}Ultrasound\\ Probe\end{tabular} & \begin{tabular}[c]{@{}c@{}}Suction\\ Instrument\end{tabular} & \begin{tabular}[c]{@{}c@{}}Clip\\ Applier\end{tabular}\\
\midrule
\multirow{6}{*}{0} & \multirow{2}{*}{3} & max & 36.90 & 30.25 & 60.44 & 23.75 & 3.57 & \textbf{19.38} & 24.24 & 28.36 \\
 &  & weighted\_mode & 36.14 & 30.89 & 58.96 & 24.38 & 4.06 & 18.04 & 24.21 & 28.10 \\
 & \multirow{2}{*}{5} & max & 36.32 & 30.19 & 59.26 & 25.66 & 5.30 & 16.45 & 22.59 & 27.97 \\
 &  & weighted\_mode & 37.58 & 30.52 & 62.06 & 23.71 & 4.13 & 18.11 & 24.72 & 28.69 \\
 & \multirow{2}{*}{7} & max & 35.70 & 29.72 & 59.33 & \textbf{27.55} & \textbf{6.13} & 16.69 & 21.21 & 28.05 \\
 &  & weighted\_mode & \textbf{38.38} & 30.96 & \textbf{64.09} & 24.70 & 4.96 & 17.48 & 23.90 & \textbf{29.21} \\
 \midrule
\multirow{6}{*}{0.5} & \multirow{2}{*}{3} & max & 36.60 & 30.82 & 57.93 & 22.86 & 3.32 & 18.47 & 25.10 & 27.87 \\
 &  & weighted\_mode & 36.64 & 30.86 & 57.70 & 22.98 & 3.32 & 19.16 & 25.10 & 27.96 \\
 & \multirow{2}{*}{5} & max & 32.09 & 25.30 & 53.72 & 19.61 & 3.71 & 18.89 & 8.92 & 23.18 \\
 &  & weighted\_mode & 32.13 & 25.37 & 54.35 & 19.52 & 3.71 & 19.27 & 8.81 & 23.31 \\
 & \multirow{2}{*}{7} & max & 36.50 & 30.94 & 58.21 & 23.43 & 3.34 & 16.15 & \textbf{25.17} & 27.68 \\
 &  & weighted\_mode & 36.76 & \textbf{31.01} & 58.31 & 22.15 & 3.34 & 16.95 & 25.02 & 27.65 \\
 \bottomrule
\end{tabular}%
}
\end{subtable}
\begin{subtable}{\textwidth}
\centering
\caption{Results on the EndoVis 2018 dataset.}
\label{tab:res2017}
\resizebox{1.0\textwidth}{!}{%
\begin{tabular}{ccccccccccc}
\toprule
\multirow{3}{*}{Threshold} & \multirow{3}{*}{\begin{tabular}[c]{@{}c@{}} Number\\ of frames\end{tabular}} & \multirow{3}{*}{\begin{tabular}[c]{@{}c@{}} Assignment\\ Strategy\end{tabular}} & \multicolumn{7}{c}{Instrument classes} & \multirow{3}{*}{\begin{tabular}[c]{@{}c@{}}mean \\ class \\ IoU\end{tabular}} \\
\cmidrule{4-10}
&  &  & \begin{tabular}[c]{@{}c@{}}Bipolar\\ Forceps\end{tabular} & \begin{tabular}[c]{@{}c@{}}Prograsp\\ Forceps\end{tabular} & \begin{tabular}[c]{@{}c@{}}Large\\ Needle Driver\end{tabular} & \begin{tabular}[c]{@{}c@{}}Monopolar\\ Curved Scissors\end{tabular} & \begin{tabular}[c]{@{}c@{}}Ultrasound\\ Probe\end{tabular} & \begin{tabular}[c]{@{}c@{}}Suction\\ Instrument\end{tabular} & \begin{tabular}[c]{@{}c@{}}Clip\\ Applier\end{tabular}\\
\midrule
\multirow{6}{*}{0} & \multirow{2}{*}{3} & max & 73.02 & 49.41 & 34.25 & 86.20 & 0.00 & 35.94 & 0.00 & 39.83 \\
 &  & weighted\_mode & 74.16 &\textbf{ 51.89} &\textbf{ 34.55} & 87.20 & 0.00 & 37.91 & 0.00 & \textbf{40.81} \\
 & \multirow{2}{*}{5} & max & 69.60 & 45.61 & 31.98 & 82.35 & 0.00 & 34.59 & 0.00 & 37.73 \\
 &  & weighted\_mode & 73.30 & 48.31 & 33.66 & 84.96 & 0.00 & 37.98 & 0.00 & 39.74 \\
 & \multirow{2}{*}{7} & max & 65.09 & 40.87 & 28.35 & 78.30 & 0.00 & 30.51 & 0.00 & 34.73 \\
 &  & weighted\_mode & 72.01 & 48.02 & 34.14 & 82.89 & 0.00 & 36.52 & 0.00 & 39.08 \\
 \midrule
\multirow{6}{*}{0.5} & \multirow{2}{*}{3} & max & 73.97 & 49.43 & 32.31 & 87.97 & 2.16 & 37.84 & 0.00 & 40.53 \\
 &  & weighted\_mode & 73.54 & 49.43 & 30.40 & 87.97 & 2.16 & 37.84 & 0.00 & 40.19 \\
 & \multirow{2}{*}{5} & max & 74.19 & 50.38 & 33.52 & 87.97 & \textbf{2.23 }& 37.23 & 0.00 & 40.79 \\
 &  & weighted\_mode & 73.76 & 48.21 & 29.50 & \textbf{88.17} & 2.16 & \textbf{38.48} & 0.00 & 40.04 \\
 & \multirow{2}{*}{7} & max &\textbf{ 74.23} & 50.82 & 33.40 & 87.37 & 2.16 & 36.76 & 0.00 & 40.68 \\
 &  & weighted\_mode & 73.96 & 48.61 & 32.17 & 88.16 & 2.16 & 37.68 & 0.00 & 40.39 \\
 \bottomrule
\end{tabular}%
}
\end{subtable}
\end{table}

\begin{figure}[h]
  \centering
  \includegraphics[width=0.9\linewidth]{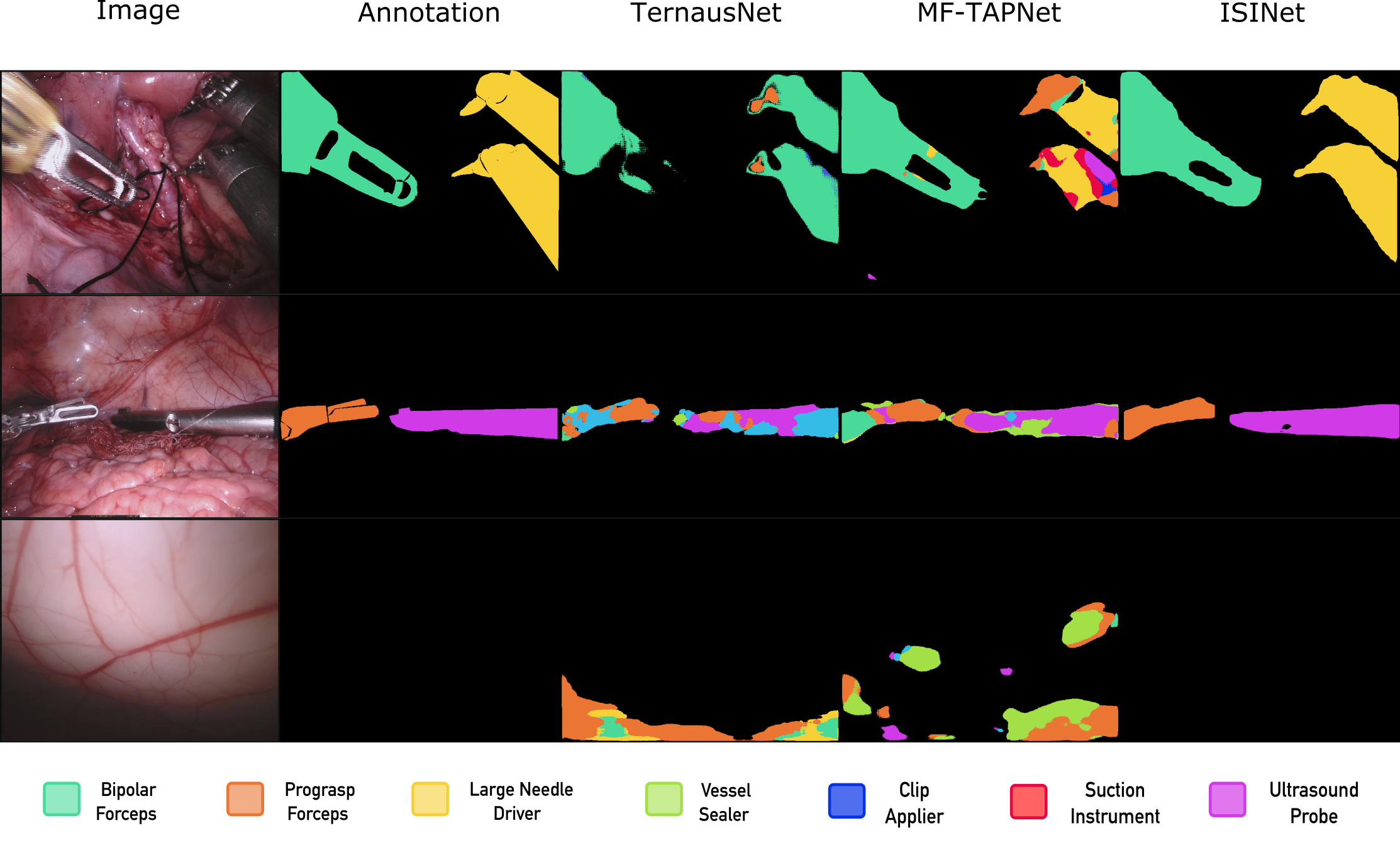}%
  \caption{Qualitative result of ISINet compared to the state-of-the-art methods TernausNet and MF-TAPNet on the EndoVis 2017 and 2018 datasets. ISINet assigns unique labels to complete instances, recovers overlooked instrument pixels and does not predict false positive instruments. Best viewed in color.}
  \label{fig:lab1}
  \centering
  \includegraphics[width=0.9\textwidth]{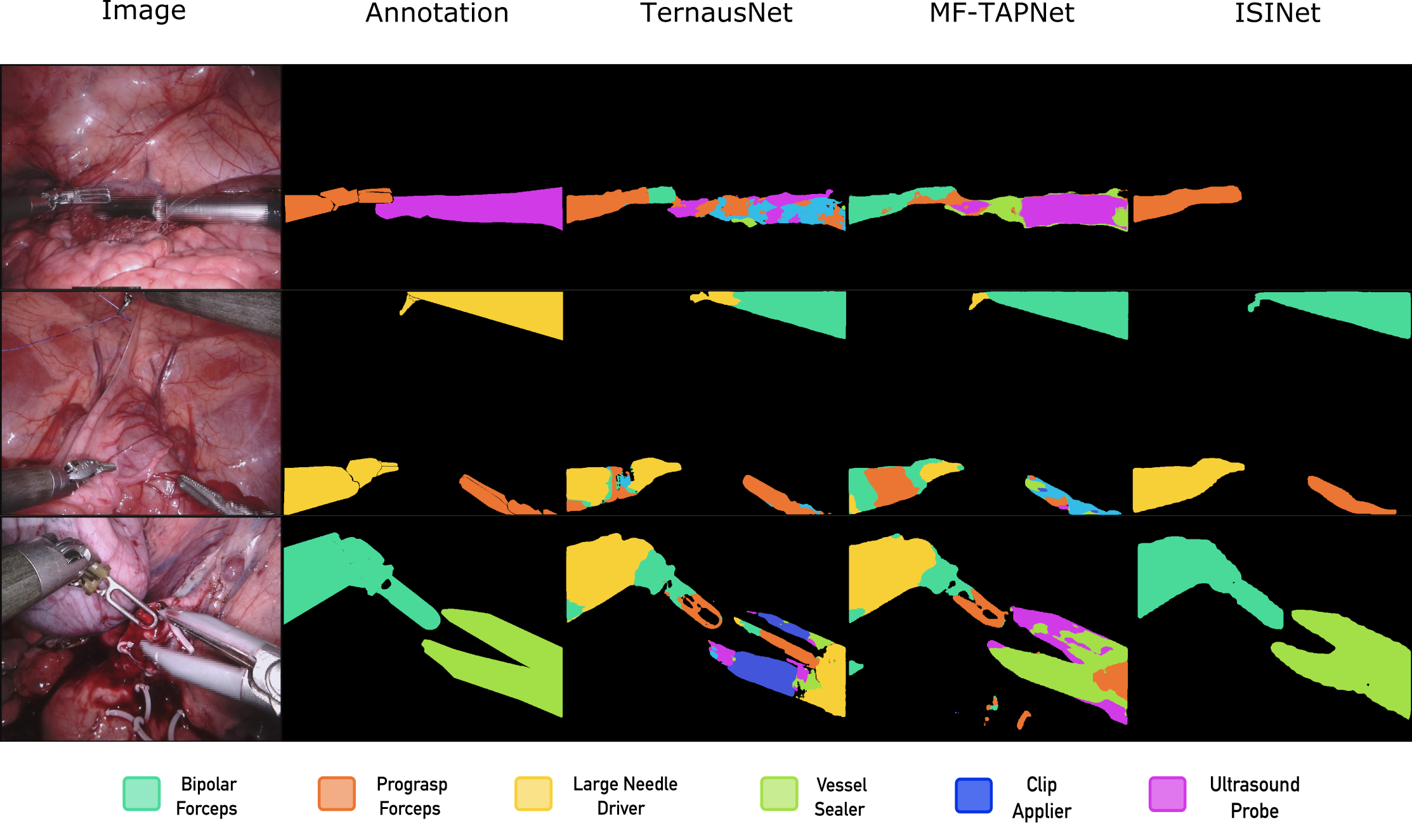}%
  \caption{Qualitative result of ISINet compared to the state-of-the-art methods TernausNet and MF-TAPNet on the EndoVis 2017 and 2018 datasets. The most common error modes of ISINet include missing instruments, mislabeling instruments and coarse segmentations. Best viewed in color.}
  \label{fig:lab2}
\end{figure}

\end{document}